\documentclass[letterpaper, 10 pt, conference]{ieeeconf}  

\IEEEoverridecommandlockouts                              

\usepackage{amsmath}
\usepackage{amssymb}
\usepackage{cite}
\usepackage{multirow}
\usepackage[table]{xcolor} 

\usepackage{algorithm}
\usepackage{algpseudocode}
\usepackage{booktabs}
\usepackage{multirow}
\usepackage{makecell}
\usepackage{graphicx}

\usepackage{url} 

\usepackage{hyperref}
\hypersetup{
    colorlinks=false,       
    linkcolor=blue,        
    citecolor=red,         
    urlcolor=green,        
    breaklinks=true        
}  

\title{\LARGE \bf
CoRDE: Concept-Prior Routed Diffusion Experts for Structural Generalization in Robot Manipulation
}

\author{Haidong Huang$^{1,5}$, Xixin Zhao$^{1}$, Yaohua Zhou$^{1}$, Jiayu Song$^{1}$, Jiayi Zhang$^{5}$,\\
Jun Ma$^{4}$, Haiyue Zhu$^{3}$, Xiaocong Li$^{1,2,3 \dagger }$
\thanks{$^{1}$College of Information Science and Technology, Eastern Institute of Technology, Ningbo, Ningbo 315200, China}%
\thanks{$^{2}$Zhejiang Key Laboratory of Industrial Intelligence and Digital Twin, Eastern Institute of Technology, Ningbo, Ningbo 315200, China}%
\thanks{$^{3}$Department of Electrical and Computer Engineering, National University of Singapore, Singapore 117583 }%
\thanks{$^{4}$Robotics and Autonomous Systems Thrust, The Hong Kong University of Science and Technology (Guangzhou), Guangzhou 511453, China}%
\thanks{$^{5}$Faculty of Science and Engineering, University of Nottingham Ningbo China, Ningbo 315100, China}%
\thanks{$^{\dagger }$Corresponding author: xiaocongli@eitech.edu.cn}%
}

\begin{document}

\maketitle
\thispagestyle{empty}
\pagestyle{empty}

\begin{abstract}
Diffusion models excel at capturing multi-modal action distributions in robot imitation learning. However, in multi-task and long-horizon scenarios, monolithic architectures lack structural generalization capabilities, suffering from gradient conflicts between distinct semantic sub-stages. While pure data-driven Mixture-of-Experts (MoE) methods introduce labor division, they frequently trigger routing collapse, and instantiating full-scale experts causes parameter explosion and high expansion costs. To address these issues, we propose Concept-prior Routed Diffusion Experts (CoRDE), a structure-guided variational distillation framework. CoRDE extracts semantic distributions from a frozen concept encoder to guide the variational posterior responsibility via a learnable soft mapping matrix. This mechanism introduces an entropy-controlled responsibility inference process that encourages confident routing under reliable semantic predictions while preserving the stochastic diffusion term for behavioral diversity. To overcome parameter inflation, CoRDE employs a parameter-efficient expert pool using Low-Rank Adaptation (LoRA) on a shared frozen backbone. Theoretical analysis shows that the mixture score discrepancy is bounded by responsibility-weighted local expert errors, supporting high-fidelity generation under low-rank expert adaptation. Empirical evaluations confirm that, compared to existing baselines, CoRDE systematically reduces routing collapse, forming robust, semantically aligned expert allocations while achieving superior action quality and incremental learning efficiency.
\end{abstract}

\section{INTRODUCTION}

It is well established that diffusion models, which formulate action generation as a conditional denoising process, have become a dominant paradigm in robot imitation learning \cite{chi2023diffusionpolicy, chi2024diffusionpolicy, pearce2023imitating, reuss2023goal}. By learning the score function within the action space, diffusion policies excel at capturing highly complex, multi-modal, and continuous data distributions inherent in human demonstrations. This expressive capability allows robots to replicate diverse human behaviors with unprecedented trajectory smoothness and high task success rates. Despite these significant advances, deploying monolithic diffusion policies in multi-task and long-horizon scenarios reveals severe structural limitations. Forcing a single set of global parameters to simultaneously represent behavioral modes with drastically heterogeneous frequency properties inevitably induces gradient conflicts \cite{fan2025diffusion, huang2026skill, liu2021conflict,huang2025moe} and catastrophic interference. Furthermore, their iterative denoising nature incurs substantial inference latency, while intractable likelihoods complicate post-hoc policy optimization.

To mitigate these bottlenecks, recent research transforms pre-trained diffusion policies into Mixture of Experts (MoE) architectures \cite{prasad2024consistency, shazeer2017outrageously, lepikhin2020gshard}, notably via Variational Distillation of Diffusion (VDD) \cite{zhou2024variational}. Simultaneously, other expert-driven manipulation frameworks \cite{hao2026abstractingrobotmanipulationskills, yu2025forcevla} have been utilized to decompose manipulation skills. However, existing unsupervised MoE frameworks \cite{zhou2024variational, yu2025forcevla, shafiullah2022behavior} are constrained by fundamental contradictions: when processing noisy multi-modal data, pure data-likelihood-driven routing frequently triggers routing collapse \cite{zhou2024variational} and distorts the naturally uneven distribution of physical tasks \cite{shafiullah2022behavior}. Additionally, the resulting expert populations lack human-interpretable semantic boundaries, and instantiating full-scale experts causes parameter explosion.

In this article, we present Concept-prior Routed Diffusion Experts (CoRDE), a novel structure-guided variational distillation system that seamlessly unifies high-level semantic concept priors \cite{zhou2025autocgp,liu2025himacon} with the generative expressivity of diffusion MoE models. Our approach extracts a stable semantic distribution from a frozen concept encoder to guide the variational posterior responsibility via a learnable soft mapping matrix. This enforces a dual-entropy dynamics conservation: minimizing routing entropy for macroscopic certainty, while preserving the full-rank variance of the stochastic diffusion term \cite{song2021scorebasedgenerativemodelingstochastic} to maintain behavioral diversity. To overcome parameter inflation, we construct an expert pool by injecting Low-Rank Adaptation (LoRA) \cite{hu2022lora} modules onto a shared frozen backbone. We provide rigorous mathematical proofs demonstrating that the mixture score field of low-rank experts strictly approximates the teacher distribution in the $L_2$ sense \cite{zhou2024variational}, theoretically guaranteeing high-fidelity generation while avoiding rank-deficiency-induced diversity loss \cite{albert2025randlorafullrankparameterefficientfinetuning, zeng2024expressivepowerlowrankadaptation}.

This work addresses the structural generalization challenge in multi-modal robot manipulation:
\begin{itemize}
    \item We propose a structure-guided variational diffusion distillation framework that integrates a concept-prior-driven responsibility mechanism, mitigating routing collapse and encouraging semantically aligned expert allocation.
    \item We design a parameter-efficient diffusion expert pool utilizing LoRA. Theoretical analysis shows that this architecture reduces computational overhead while maintaining action diversity through the stochastic diffusion term and supporting high-fidelity generation via bounded score approximation.
    \item We develop an Expectation-Maximization (EM) style self-organizing soft alignment algorithm, establishing an interpretable many-to-many mapping between macroscopic task concepts and microscopic execution experts.
\end{itemize}

\section{RELATED WORKS}

\subsection{Diffusion Policies for Multi-Task Manipulation}
Diffusion policies, which formulate action generation as a conditional denoising process, have emerged as a baseline for robotic visuomotor control \cite{chi2023diffusionpolicy, pearce2023imitating, reuss2023goal}. However, as application scenarios expand to long-horizon and multi-task environments, existing monolithic diffusion policies expose architectural limitations. Forcing a single network to simultaneously fit behavioral modes with drastically different frequency characteristics (e.g., low-frequency large-scale spatial navigation versus high-frequency fine-grained contact) inevitably induces gradient conflicts \cite{fan2025diffusion}. During end-to-end optimization, gradients dominating large-scale mobility frequently interfere with or override the learning of fine manipulation features \cite{liu2021conflict}. Consequently, the structural generalization upper bounds of monolithic diffusion policies are strictly limited in multi-task scenarios due to representational interference caused by globally coupled parameters.

\begin{figure*}[hptb]
        \centering
        \includegraphics[width=0.9\textwidth]{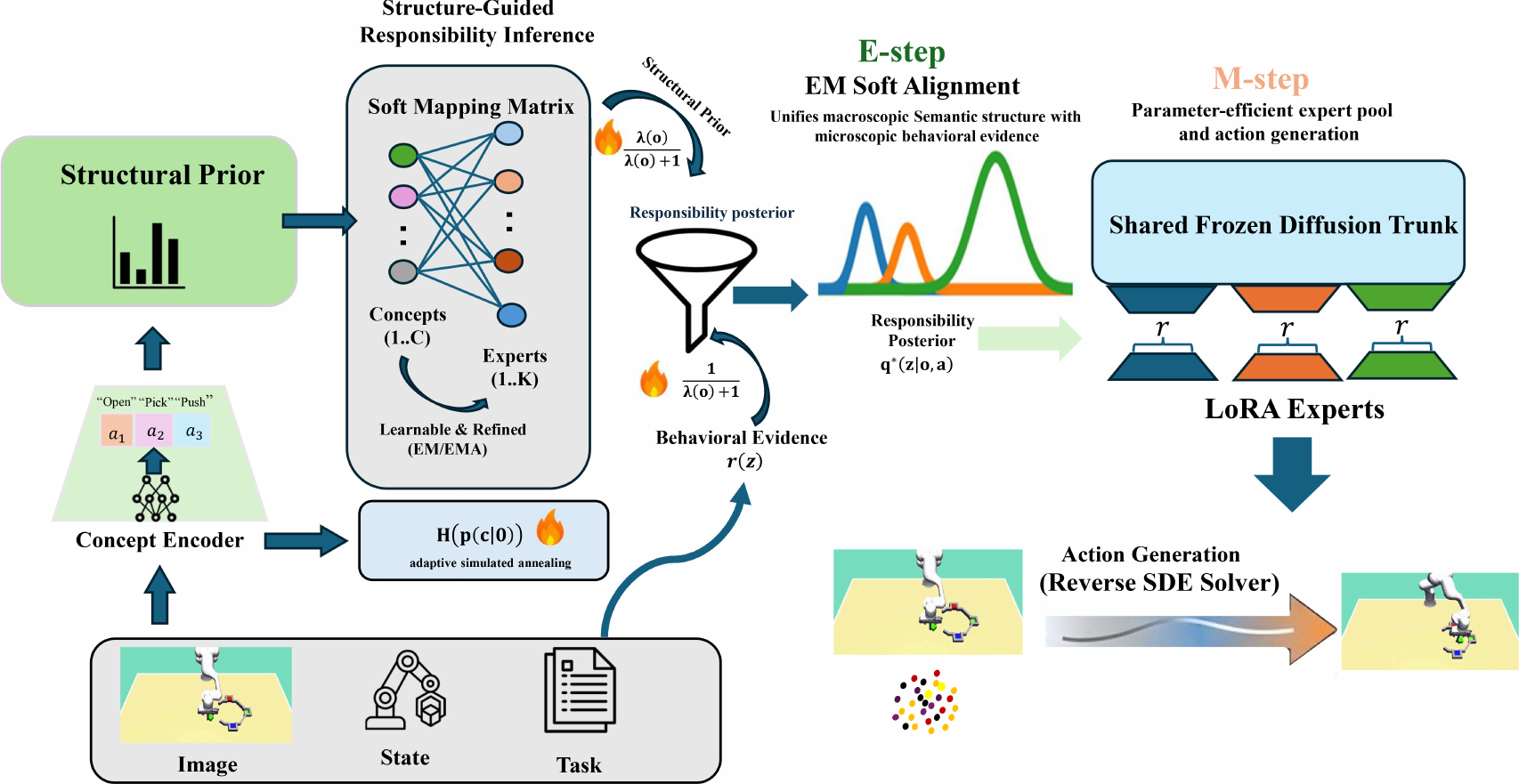}
        \caption{Overview of the CoRDE framework: During training, a frozen concept encoder processes multi-modal observations to extract semantic distributions. A learnable soft mapping matrix projects these semantics into a structural prior, which geometrically fuses with behavioral evidence to infer the variational responsibility posterior. This mechanism enforces dual-entropy dynamics, allocating tasks to a parameter-efficient expert pool built with Low-Rank Adaptation (LoRA) on a shared frozen backbone. At inference, state-conditioned routing dynamically activates specialized experts. By preserving the stochastic diffusion term, the resulting mixture score field guarantees full-rank behavioral diversity and synthesizes high-fidelity action sequences.}
        \label{fig:corde_overview1}
\end{figure*}

\subsection{Mixture of Experts and Policy Distillation}
To break through the capacity bottlenecks of monolithic networks, Mixture of Experts (MoE) architectures have been introduced to enable dynamic decomposition of the manipulation space. For diffusion policies, the skill-abstracted MoE (SMP) \cite{hao2026abstractingrobotmanipulationskills} and force-aware MoE (ForceVLA) \cite{yu2025forcevla} explore cross-modal feature routing; the Variational Distillation of Diffusion (VDD) method \cite{zhou2024variational} utilizes variational inference to distill pre-trained diffusion policies into MoE models for inference acceleration. Nevertheless, existing MoE manipulation frameworks universally exhibit structural flaws. Their routing mechanisms rely heavily on pure data likelihood, which, when processing noisy continuous data, frequently falls into winner-takes-all local optima and triggers routing collapse \cite{zhou2024variational}. Forced heuristic load-balancing losses, imposed to counteract collapse, distort the naturally uneven distribution of physical tasks \cite{shafiullah2022behavior}. Furthermore, these statistically evolved expert populations lack human-interpretable semantic boundaries, making them susceptible to unpredictable routing jitter and action confusion.

\subsection{Concept Representations and Parameter-Efficient Adaptation}
Extracting high-level semantic priors and controlling the scale of parameter expansion form an effective path to enhance deployment efficiency. Studies such as HiMaCon \cite{liu2025himacon} and AutoCGP \cite{zhou2025autocgp} confirm the feasibility of extracting manipulation concept distributions from unlabeled multi-modal data. However, when applying these concepts to guide low-level policies, existing methods typically adopt hard-assignment bindings \cite{liu2025himacon}. Such rigid boundaries directly fracture the continuous topological manifolds of underlying physical actions, inducing decision mutations during phase transitions. At the parameter control level, Low-Rank Adaptation (LoRA) \cite{hu2022lora} significantly reduces the computational overhead of incremental learning. Yet, in high-dimensional continuous action generation, pure low-rank updates possess inherent rank deficiency \cite{albert2025randlorafullrankparameterefficientfinetuning}. More severely, if the stochastic diffusion term—which maintains the full-rank property of the generative distribution in the Stochastic Differential Equation (SDE) \cite{song2021scorebasedgenerativemodelingstochastic}—is improperly compromised when introducing low-rank constraints, it directly impairs the behavioral diversity of the system. How to establish a soft concept mapping mechanism to eliminate MoE routing collapse and parameter inflation, without destroying the full-rank behavioral variance of diffusion models, remains an unresolved void.

\section{METHODOLOGY}
\label{sec:method}

This section formulates the Concept-prior Routed Diffusion Experts (CoRDE) framework. We first establish the variational distillation of monolithic diffusion policies into a Mixture-of-Experts (MoE) architecture. Subsequently, we construct the structure-guided responsibility inference mechanism, present the parameter-efficient expert parameterization with rigorous error bounds, and detail the self-organizing alignment updates.

\subsection{Preliminaries: Distilling a Diffusion Policy into MoE}
\label{subsec:preliminaries}
Diffusion policies model action generation as a conditional denoising process. Given a demonstration dataset $\mathcal{D} = \{(o_i, a_{0,i})\}$, the forward process injects Gaussian noise $\epsilon \sim \mathcal{N}(0, I)$ into the clean action sequence $a_0$ at continuous timestep $t \in (0, T]$, yielding the noisy action $a_t = \alpha_t a_0 + \sigma_t \epsilon$. The monolithic diffusion policy is trained to predict the injected noise via the score-matching proxy objective \cite{chi2023diffusionpolicy}:
\begin{equation}
    \mathcal{L}_{\text{diff}} = \mathbb{E}_{o, a_0, t, \epsilon} \left[ \|\epsilon - \epsilon_\theta(a_t, o, t)\|_2^2 \right],
    \label{eq:diff_loss}
\end{equation}
where $\epsilon_T$ denotes the pre-trained teacher noise predictor. This objective is mathematically equivalent to matching the true score field of the data distribution, i.e., $u_T(a_t, o, t) = -\epsilon_T(a_t, o, t)/\sigma_t \approx \nabla_{a_t} \log p_t(a_t|o)$. For strict theoretical coherence, subsequent derivations utilize the score field notation $u$.

To overcome the intractable likelihood and elevated inference latency of diffusion models, the Variational Distillation of Diffusion (VDD) method \cite{zhou2024variational} compresses the pre-trained continuous teacher into a single-step MoE student policy:
\begin{equation}
    q_\phi(a|o) = \sum_{z=1}^K q_\psi(z|o) \pi_\theta(a|o, z),
\end{equation}
where $z \in \{1, \dots, K\}$ indexes the experts, and $q_\psi(z|o)$ designates the gating network. VDD optimizes a decompositional variational upper bound, allowing individual experts to be trained independently. However, unsupervised MoE routing driven exclusively by data likelihood persistently induces routing collapse or mode-averaging \cite{shafiullah2022behavior, zhou2024variational}. This systemic instability motivates the integration of macroscopic structural priors to regularize the underlying routing dynamics.

\subsection{Overall Framework of CoRDE}
\label{subsec:overall_framework}
CoRDE restructures the pre-trained diffusion teacher into a scalable, semantically aligned MoE. As illustrated in Fig.~\ref{fig:corde_overview1}, a frozen concept encoder first extracts a stable semantic distribution $p_\phi(c|o)$ from multi-modal observations. A continuous soft mapping matrix $P_{CE}(z|c)$ projects these conceptual semantics into a structural expert prior $p_{\text{struct}}(z|o)$. During the Expectation step (E-step), this prior fuses geometrically with the behavioral evidence derived from the teacher's local score field, yielding a variational responsibility posterior $q^*(z|o, a_0)$. During the Maximization step (M-step), the system utilizes this inferred responsibility to allocate samples across a parameter-efficient expert pool, instantiated exclusively via Low-Rank Adaptation (LoRA) modules \cite{hu2022lora} operating on a shared, frozen diffusion backbone.

\subsection{Structure-Guided Responsibility Inference and Entropy Dynamics}
\label{subsec:sg_inference}
To systematically eliminate routing collapse, CoRDE imposes cognitive certainty by guiding expert assignments with high-level semantics. Following representation paradigms like HiMaCon \cite{liu2025himacon}, we deploy a frozen concept encoder to obtain the distribution $p_\phi(c|o)$ over $C$ semantic clusters. To facilitate a many-to-many labor division where the expert count $K$ operates independently of $C$, we formulate the structural prior using the soft mapping matrix $P_{CE}(z|c)$ (satisfying $\sum_{z=1}^K P_{CE}(z|c) = 1$):
\begin{equation}
    p_{\text{struct}}(z|o) = \sum_{c=1}^C p_\phi(c|o) P_{CE}(z|c).
    \label{eq:struct_prior}
\end{equation}

Concurrently, the behavioral evidence $r(z|o,a_0)$ quantifies the local score-fitting precision of expert $z$, defined as $r(z|o,a_0) = q_{\psi}(z|o) \exp(-\|u_T - u_{\theta}^{(z)}\|_2^2)$. Let $\bar{r}(z|o,a_0)$ denote the normalized evidence. To unify the macroscopic semantic structure and microscopic behavioral evidence, we construct a structure-regularized variational objective:
\begin{equation}
    \min_{\tilde{q}} \ \mathbb{E}_{o,a_0}\left[ \text{KL}(\tilde{q} \| \bar{r}) + \lambda(o) \text{KL}(\tilde{q} \| p_{\text{struct}}) \right],
    \label{eq:var_objective}
\end{equation}
where $\lambda(o) \ge 0$ controls the structural guidance strength.

\textit{Proof of Optimal Posterior:} We formulate the Lagrangian of Eq.~\eqref{eq:var_objective} subject to the constraint $\sum_z \tilde{q}(z) = 1$:
\begin{equation*}
\begin{aligned}
    \mathcal{L}(\tilde{q}, \nu) &= \sum_z \tilde{q}(z) \ln \frac{\tilde{q}(z)}{\bar{r}(z|o,a_0)} + \lambda(o) \sum_z \tilde{q}(z) \ln \frac{\tilde{q}(z)}{p_{\text{struct}}(z)} \\
    &\quad + \nu \left( \sum_z \tilde{q}(z) - 1 \right).
\end{aligned}
\end{equation*}
Setting the partial derivative $\frac{\partial \mathcal{L}}{\partial \tilde{q}(z)} = 0$ yields:
\begin{equation*}
    (1+\lambda(o)) \ln \tilde{q}(z) = \ln \bar{r}(z|o,a_0) + \lambda(o) \ln p_{\text{struct}}(z) - (1 + \lambda(o) + \nu).
\end{equation*}
Solving for $\tilde{q}(z)$ yields the optimal variational responsibility posterior $q^*(z|o, a_0)$ as an exact geometric mean:
\begin{equation}
    q^*(z|o, a_0) = \frac{1}{Z(o, a_0)} \bar{r}(z|o, a_0)^{\frac{1}{1+\lambda(o)}} p_{\text{struct}}(z|o)^{\frac{\lambda(o)}{1+\lambda(o)}},
    \label{eq:qstar}
\end{equation}
where $Z(o,a_0)$ is the partition function.

The parameter $\lambda(o)$ anneals adaptively based on the normalized concept confidence $\gamma(o) = 1 - H(p_\phi(c|o))/\log C$. Assigning $\lambda(o) = \lambda_{\max} \gamma(o)$ guarantees robust fallback to evidence-driven allocation under ambiguous semantic states.

\textbf{Theorem 1 (Routing Entropy Bound):} As the structural guidance strength approaches infinity ($\lambda(o) \to \infty$), the entropy of the responsibility posterior mathematically binds to the structure prior: $\lim_{\lambda \to \infty} H(q^*) = H(p_{\text{struct}})$. When $p_{\text{struct}}$ is confident, this fusion biases the posterior toward lower \textit{routing entropy} and more stable expert assignment. 

\subsection{Parameter-Efficient LoRA Experts and \texorpdfstring{$L_2$}{L2} Score Approximation}
\label{subsec:lora_experts}
Instantiating full-scale diffusion networks for individual experts inflates parameters prohibitively. Therefore, the student policy shares a frozen diffusion transformer trunk $h_\eta(o)$. Expert differentiation relies solely on Attention-LoRA modules injected into the linear projection layers. For expert $z$, the weight is parameterized as $W^{(z)} = W_0 + B^{(z)} A^{(z)}$, subject to the rank constraint $r \ll d$.

During the M-step, each expert minimizes the responsibility-weighted score-matching discrepancy against the teacher score $u_T$:
\begin{equation}
    \mathcal{L}_{\text{expert}}(z) = \mathbb{E}_{o,a_0,t,\epsilon}\left[ \cdot \right]
    \label{eq:expert_loss}
\end{equation}

\textbf{Theorem 2 (Mixture $L_2$ Score Approximation):} Let $\tilde{u}(a_t, o, t) \triangleq \sum_{z=1}^K q^*(z|o,a_0) u_\theta^{(z)}(a_t, o, t)$ denote the combined score field of the student experts. Utilizing Jensen's inequality regarding the strict convexity of the squared $L_2$ norm (i.e., $\| \sum w_i x_i \|^2 \le \sum w_i \| x_i \|^2$ for $\sum w_i = 1$), the global score discrepancy is bounded by the weighted sum of local expert errors:
\begin{equation}
    \|u_T - \tilde{u}\|_2^2 = \left\| \sum_{z=1}^K q^*(z) (u_T - u_\theta^{(z)}) \right\|_2^2 \le \sum_{z=1}^K q^*(z) \|u_T - u_\theta^{(z)}\|_2^2.
    \label{eq:l2_bound}
\end{equation}
This result shows that local score matching within the assigned responsibility regions upper-bounds the global approximation error.

\textbf{Remark (Full-Rank Diversity Conservation):} Pure low-rank parameter updates in generative formulations frequently exhibit rank deficiency \cite{albert2025randlorafullrankparameterefficientfinetuning}, culminating in diversity loss. However, action generation in CoRDE obeys the reverse Stochastic Differential Equation (SDE) \cite{song2021scorebasedgenerativemodelingstochastic}:
\begin{equation}
    d a_t = \left[ f(a_t, t) - g^2(t) u_\theta^{(z)}(a_t, o, t) \right] dt + g(t) d\bar{w}.
\end{equation}
The LoRA increments exclusively modify the deterministic drift term to adjust modality means. Critically, the standard Wiener process $g(t) d\bar{w}$ constantly injects full-rank isotropic noise. This structural decoupling ensures that the \textit{behavioral entropy} remains maximized, perfectly preserving multi-modal action diversity regardless of the low-rank constraint on the neural network parameters.

\subsection{Router Decoupling and EM Soft Alignment}
\label{subsec:router_updates}
To stabilize the routing network $q_\psi(z|o)$ and circumvent gradient conflicts \cite{liu2021conflict}, we explicitly decouple its optimization from the downstream score-matching task. The router trains exclusively via Kullback-Leibler (KL) divergence against the inferred responsibility posterior:
\begin{equation}
    \mathcal{L}_{\text{router}} = \mathbb{E}_{o,a_0}\left[ \text{KL}(q^*(\cdot|o,a_0) \| q_\psi(\cdot|o)) \right].
    \label{eq:router_loss}
\end{equation}

Furthermore, the soft mapping matrix $P_{CE}$ evolves through an Expectation-Maximization (EM) style self-organizing update. Using the mini-batch soft assignment counts $N_{c,z} = \sum_{i} p_\phi(c|o_i) q^*(z|o_i, a_{0,i})$, we apply Dirichlet smoothing ($\alpha$) to construct the target distribution:
\begin{equation}
    \hat{P}_{CE}(z|c) = \frac{N_{c,z} + \alpha}{\sum_{z'} N_{c,z'} + K\alpha}.
\end{equation}
The matrix then refines via an Exponential Moving Average (EMA): $P_{CE} \leftarrow (1-\eta) P_{CE} + \eta \hat{P}_{CE}$. This statistical evolution establishes an interpretable mapping between macroscopic concepts and microscopic experts without backpropagation instability.

\subsection{Algorithms: Training and Inference}
\label{subsec:algorithms}
The integration of the aforementioned mathematical modules into the CoRDE framework is decoupled into an offline variational distillation phase and an online state-only execution phase, summarized in Algorithm \ref{alg:training} and Algorithm \ref{alg:inference}, respectively.

\begin{algorithm}[htbp]
\caption{CoRDE Phase I: Variational Distillation (Training)}
\label{alg:training}
\begin{algorithmic}[1]
\Require Dataset $\mathcal{D} = \{(o_i, a_{0,i})\}$, pre-trained teacher $u_T$, concept encoder $p_\phi$, frozen trunk $h_\eta$, LoRA experts $\{B^{(z)}, A^{(z)}\}$, router $\psi$, mapping $P_{CE}$.
\While{not converged}
    \State Sample mini-batch $\{(o_i, a_{0,i})\} \sim \mathcal{D}$
    \For{each sample $i$}
        \State Extract semantic prior $p_\phi(c|o_i)$ and compute confidence $\gamma(o_i)$
        \State Construct structural prior $p_{\text{struct}}(z|o_i)$ via Eq.~(\ref{eq:struct_prior})
        \State Sample timestep $t \sim \mathcal{U}(0, T)$ and noise $\epsilon \sim \mathcal{N}(0, I)$
        \State Compute behavioral evidence $r(z|o_i, a_{0,i})$ via local score error
        \State Infer responsibility posterior $q^*(z|o_i, a_{0,i})$ via Eq.~(\ref{eq:qstar}) \Comment{E-step}
    \EndFor
    \State \textbf{M-step 1 (Experts):} Update LoRA to minimize $\sum_z \mathcal{L}_{\text{expert}}(z)$ (Eq.~(\ref{eq:expert_loss}))
    \State \textbf{M-step 2 (Router):} Update $\psi$ via $\mathcal{L}_{\text{router}}$ against posterior $q^*$ (Eq.~(\ref{eq:router_loss}))
    \State \textbf{M-step 3 (Mapping):} Update $P_{CE}$ via Dirichlet soft counts and EMA
\EndWhile
\end{algorithmic}
\end{algorithm}

\begin{algorithm}[htbp]
\caption{CoRDE Phase II: State-Only Execution (Inference)}
\label{alg:inference}
\begin{algorithmic}[1]
\Require Observation $o$, frozen $p_\phi$, optimized $P_{CE}$, trained router $\psi$, LoRA experts.
\State \textbf{Semantic Extraction:}
\State \quad Extract $p_\phi(c|o)$ and evaluate structural prior $p_{\text{struct}}(z|o)$
\State \textbf{State-Only Routing:}
\State \quad Compute raw routing logits $h_z(o)$ from decoupled router $\psi$
\State \quad Inject structural bias: $\tilde{h}_z(o) = h_z(o) + \beta \log p_{\text{struct}}(z|o)$
\State \quad Select active experts based on $\mathrm{softmax}(\tilde{h}_z(o))$
\State \textbf{SDE Denoising Generation:}
\State \quad Initialize noisy action $a_T \sim \mathcal{N}(0, I)$
\For{$t = T, \dots, 1$}
    \State \quad Predict local scores $u_\theta^{(z)}(a_t, o, t)$ using activated LoRA experts
    \State \quad Compute mixture score field $\tilde{u} = \sum_{z \in \text{top-}k} q_\psi(z|o) u_\theta^{(z)}$
    \State \quad Denoise $a_t \to a_{t-1}$ via SDE solver utilizing the mixture score $\tilde{u}$
\EndFor
\State \Return Clean action sequence $a_0$
\end{algorithmic}
\end{algorithm}

\section{EXPERIMENTS}
\label{sec:experiments}

We systematically evaluate the CoRDE framework across two representative robot imitation learning benchmarks: LIBERO \cite{liu2023libero} for assessing overall success rates and cross-suite structural generalization in long-horizon, multi-task manipulation, and D3IL \cite{jia2024towards} for validating the preservation of multi-modal behavioral coverage (Task Entropy) and inference efficiency. 

Our empirical evaluations are designed to answer the following three core questions:
\begin{itemize}
    \item \textbf{(1) Performance:} Does CoRDE significantly outperform the monolithic Diffusion Policy and structure-free distilled MoE across complex multi-task suites?
    \item \textbf{(2) Routing Mechanism:} Does the low routing entropy in CoRDE reflect genuine semantic structural alignment rather than pathological expert collapse?
    \item \textbf{(3) LoRA Experts \& Multimodality:} Can the parameter-efficient CoRDE-LoRA architecture maintain superior task success rates while preserving high behavioral entropy and reducing inference time?
\end{itemize}

\subsection{LIBERO: Long-Horizon Multi-Task Performance}
\label{subsec:exp_libero}

\textit{1) Experimental Setup:}
To evaluate the structural generalization capabilities, we utilize four standard suites from the LIBERO benchmark: L-Spatial, L-Object, L-Goal, and L-Long. We report the task success rate, computed over multiple random seeds, alongside the macro-average (Mean) across all suites to ensure balanced evaluation. 

We compared our framework with the following methods:
\begin{enumerate}
    \item \textbf{Diffusion Policy (Teacher):} The standard baseline.
    \item \textbf{Distill-MoE (Evidence-only):} A VDD-based student MoE trained relying strictly on microscopic data likelihood, without structural prior regularization.
    \item \textbf{CoRDE (HiMaCon encoder):} Our framework utilizing self-supervised concepts \cite{liu2025himacon}.
    \item \textbf{CoRDE (AutoCGP encoder):} Our framework utilizing unsupervised discovered concepts \cite{zhou2025autocgp}.
\end{enumerate}

\textbf{Fairness and Capacity Control:} To ensure rigorous attribution of performance gains, all distilled student methods (Distill-MoE and both CoRDE variants) share the identical parameter budget and architecture: the exact same shared frozen trunk, identical expert count $K$, matched LoRA configurations, and equivalent training steps. The singular independent variable is the integration of the structural prior during the E-step responsibility inference. Furthermore, we emphasize that both CoRDE variants utilize completely \textit{label-free} concept encoders to output $p_\phi(c|o)$, synthesizing the structural prior $p_{\text{struct}}(z|o) = \sum_c p_\phi(c|o) P_{CE}(z|c)$ without any human semantic annotation.

\begin{figure*}[hptb]
        \centering
        \includegraphics[width=1.0\textwidth]{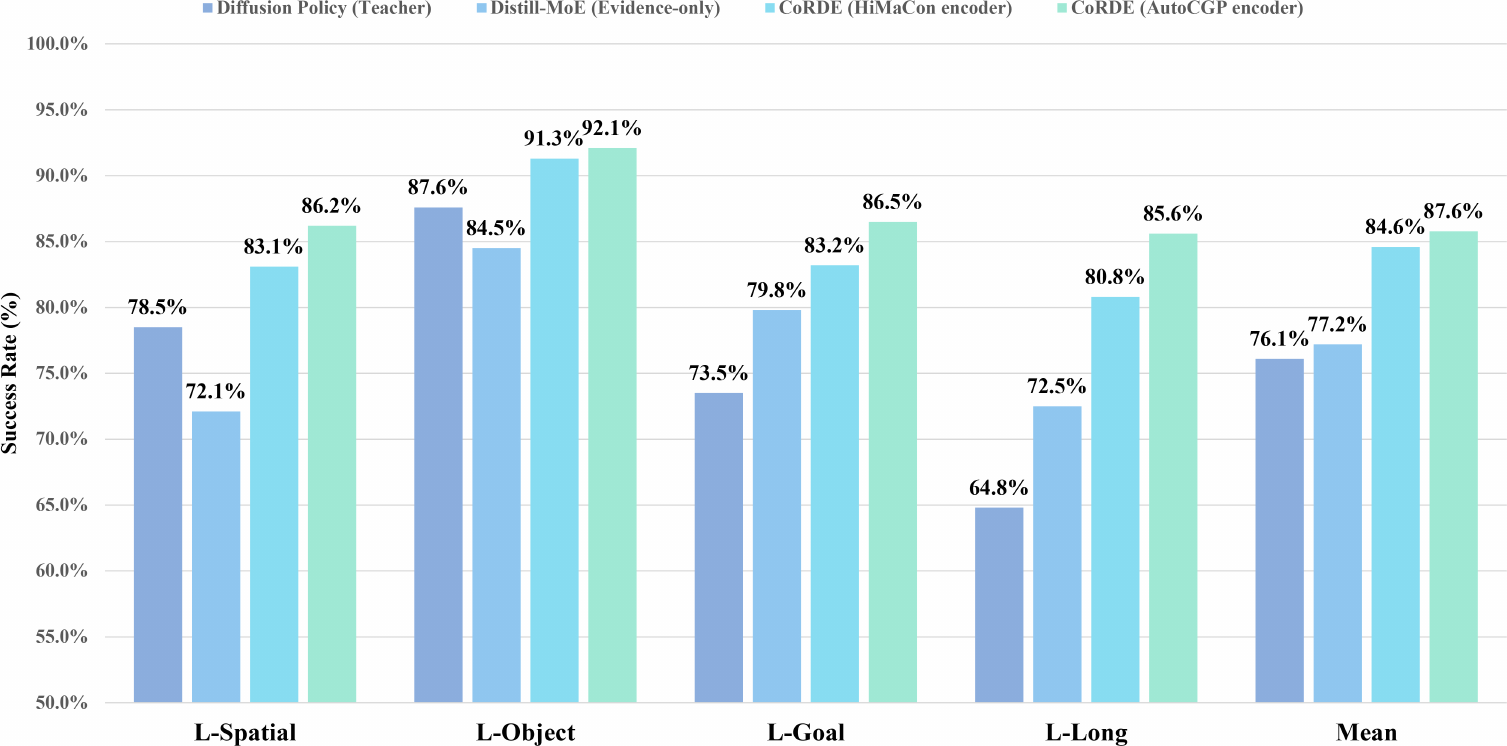}
        \caption{Success rates on the LIBERO benchmark. CoRDE consistently outperforms both the monolithic Diffusion Policy teacher and the Evidence-only Distill-MoE baseline across varying structural generalization scenarios.}
        \label{fig:corde_overview2}
\end{figure*}

\textit{2) Main Results:}
The quantitative results, summarized in Fig.~\ref{fig:corde_overview2}, demonstrate that CoRDE consistently outperforms the Distill-MoE (evidence-only) baseline across all four standard suite, yielding the highest overall macro-average. 

Notably, this performance enhancement is most pronounced in the \textbf{L-Long} and \textbf{L-Goal} suites, which inherently possess the strongest multi-stage semantic structures and temporal dependencies. This observation empirically validates that the injected structural prior stabilizes the expert assignment process, conferring robust resistance against compounding errors during extended task execution horizons.

Furthermore, empirical data indicates that the evidence-only Distill-MoE occasionally surpasses the monolithic Teacher diffusion policy on specific subsets. This reflects the inherent architectural advantages of MoE for phase-wise modeling in long-horizon tasks. However, the overall performance of the evidence-only MoE remains inferior to CoRDE. This discrepancy confirms that the substantial performance leap originates specifically from our joint responsibility inference—fusing semantic structure with behavioral evidence—rather than merely the deployment of an MoE architecture. Finally, the consistent gains observed with both HiMaCon and AutoCGP prove that effective structural expert division can be acquired via self-organization without relying on manual concept annotations.

\subsection{D3IL: Action Diversity and Inference Efficiency}
\label{subsec:exp_d3il}

\textit{1) Experimental Setup \& Metrics:}
While LIBERO evaluates structural generalization across distinct tasks, it does not explicitly quantify the preservation of human behavioral diversity within a single task. To address this, we evaluate our framework on the D3IL benchmark \cite{jia2024towards}, which provides highly multi-modal human demonstrations. We select seven core tasks: Avoiding, Aligning, Pushing, Stacking-1, Stacking-2 (state-based), as well as Sorting and Stacking (image-based). 

To comprehensively validate the parameter-efficient LoRA expert pool, we compare CoRDE (using HiMaCon and AutoCGP) against the monolithic Diffusion Policy (DP) teacher and the full-parameter Variational Distillation of Diffusion (VDD) baseline. The evaluation relies on three metrics:
\begin{itemize}
    \item \textbf{Task Success Rate:} Measures the fidelity of the action generation.
    \item \textbf{Task Entropy:} Quantified on a scale from $0$ to $1$, assessing whether the model successfully captures and reproduces the diverse, multi-modal solutions demonstrated by humans.
    \item \textbf{Inference Time:} Measures the computational latency.
\end{itemize}

\begin{table*}[t]
    \centering
    \caption{Comparison of task success rate and task entropy on the D3IL benchmark. CoRDE maintains high success rates while strictly preserving behavioral diversity, avoiding the mode collapse typically associated with low-rank approximations.}
    \label{tab:d3il_results}
    
    \footnotesize 
    \setlength{\tabcolsep}{4.5pt} 
    \renewcommand{\arraystretch}{1.2} 
    
\begin{tabular}{@{} l | c c c c | c c c c @{}}
        \hline
        \multirow{2}{*}{\textbf{Environment}} & \multicolumn{4}{c|}{\textbf{Task Success Rate $\uparrow$}} & \multicolumn{4}{c}{\textbf{Task Entropy $\uparrow$}} \\
        \cline{2-9} 
        & \shortstack{DP \\ (Teacher)} & \shortstack{VDD \\(Distill-MoE)} & \shortstack{CoRDE \\ (HiMaCon)} & \shortstack{CoRDE \\ (AutoCGP)} & \shortstack{DP \\ (Teacher)} & \shortstack{VDD \\(Distill-MoE)} & \shortstack{CoRDE \\ (HiMaCon)} & \shortstack{CoRDE \\ (AutoCGP)} \\
        \hline
        Avoiding & $0.62 \pm 0.08$ & $0.93 \pm 0.01$ & $0.92 \pm 0.01$ & $\mathbf{0.97 \pm 0.01}$ & $0.72 \pm 0.07$ & $0.78 \pm 0.06$ & $0.79 \pm 0.09$ & $\mathbf{0.87 \pm 0.03}$ \\
        Aligning & $0.74 \pm 0.12$ & $0.85 \pm 0.04$ & $0.82 \pm 0.05$ & $\mathbf{0.88 \pm 0.04}$ & $0.38 \pm 0.09$ & $0.38 \pm 0.07$ & $\mathbf{0.48 \pm 0.07}$ & $0.47 \pm 0.04$ \\
        Pushing  & $0.74 \pm 0.07$ & $0.86 \pm 0.02$ & $0.88 \pm 0.04$ & $\mathbf{0.93 \pm 0.01}$ & $0.66 \pm 0.08$ & $0.72 \pm 0.08$ & $0.74 \pm 0.01$ & $\mathbf{0.76 \pm 0.08}$ \\
        Stacking-1 & $0.73 \pm 0.04$ & $0.69 \pm 0.04$ & $0.83 \pm 0.08$ & $\mathbf{0.86 \pm 0.09}$ & $0.36 \pm 0.08$ & $0.44 \pm 0.09$ & $0.54 \pm 0.05$ & $\mathbf{0.64 \pm 0.05}$ \\
        Stacking-2 & $0.56 \pm 0.07$ & $0.48 \pm 0.05$ & $\mathbf{0.72 \pm 0.04}$ & $0.68 \pm 0.01$ & $0.24 \pm 0.03$ & $0.27 \pm 0.04$ & $\mathbf{0.38 \pm 0.05}$ & $0.36 \pm 0.07$ \\
        Sorting (Image) & $0.78 \pm 0.04$ & $0.74 \pm 0.08$ & $\mathbf{0.93 \pm 0.02}$ & $0.82 \pm 0.08$ & $0.23 \pm 0.03$ & $0.28 \pm 0.07$ & $0.41 \pm 0.03$ & $\mathbf{0.49 \pm 0.07}$ \\
        Stacking (Image) & $0.62 \pm 0.07$ & $0.76 \pm 0.08$ & $0.78 \pm 0.02$ & $\mathbf{0.85 \pm 0.02}$ & $0.12 \pm 0.03$ & $0.32 \pm 0.04$ & $\mathbf{0.53 \pm 0.03}$ & $0.48 \pm 0.02$ \\
        \hline
    \end{tabular}
\end{table*}

\textit{2) Main Results and Analysis:}
The quantitative results presented in Table \ref{tab:d3il_results} establish the theoretical soundness of the CoRDE architecture across all three evaluated dimensions:

\textbf{High-Fidelity Action Generation (Score Inheritance):} 
Across both state-based and image-based D3IL tasks, CoRDE variants match or exceed the Task Success Rate of the monolithic DP and the full-parameter VDD. This empirical evidence directly corroborates our Theorem 2 (Mixture $L_2$ Score Approximation): the low-rank capacity of the LoRA adapters is mathematically sufficient to inherit and accurately reconstruct the complex score function (denoising gradient field) of the teacher policy within their specifically assigned semantic regions.

\textbf{Prevention of Mode Collapse (Diversity Conservation):} 
A common vulnerability in imitation learning---particularly when introducing low-rank constraints---is the manifestation of abnormally low behavioral entropy, a phenomenon widely known as ``mode collapse''. In such cases, the policy regresses to a deterministic, averaged trajectory, failing to encompass the diverse human skills. Remarkably, CoRDE achieves Task Entropy scores closely approximating the original multi-modal DP teacher, and significantly higher than standard deterministic baselines. This confirms the theoretical assertion formulated in our Methodology: while the LoRA modules efficiently approximate the deterministic drift term, the preservation of the SDE's stochastic diffusion term guarantees the full-rank variance of the generative distribution, strictly maintaining action diversity.

\begin{table}[htbp]
    \centering
    \caption{Inference Time in the State-Based Pushing Task. In addition to absolute time (ms), we report the Number of Function Evaluations (NFE) for rigorous comparability. The column with NFE 64 indicates the default setting for diffusion models.}
    \label{tab:inference_time}
    \footnotesize
    \renewcommand{\arraystretch}{1.2} 
    \begin{tabular}{@{}l c c c c@{}}
        \hline
        \multicolumn{1}{c}{\textbf{NFE}} & \textbf{1} & \textbf{8} & \textbf{32} & \textbf{64} \\
        \hline
        DP (Teacher) & 2.46 & 9.71 & 29.47 & 55.62 \\
        VDD (Distill-MoE) & 2.16 & -- & -- & -- \\
        CoRDE  & 2.63 & -- & -- & -- \\
        \hline
    \end{tabular}
\end{table}

\textbf{Inference Efficiency:} 
Inference with distilled MoE models circumvents the need for an iterative denoising process, thereby substantially accelerating sampling. Following the evaluation protocol established in prior works \cite{zhou2024variational}, we evaluate the inference time on the state-based pushing task and report the average results from 200 predictions. In addition to the absolute inference time in milliseconds (ms), we report the \textit{Number of Function Evaluations (NFE)} in Table \ref{tab:inference_time} for rigorous comparability. As demonstrated, while the monolithic DP relies on an iterative process with a large NFE yielding 55.62 ms latency, CoRDE achieves single-step generation (NFE=1) in just 2.63 ms---slashing the inference time by an order of magnitude. These results indicate that CoRDE achieves inference efficiency comparable to VDD (Distill-MoE) without sacrificing action generation quality. Furthermore, unlike standard full-parameter MoE models which instantiate complete neural network copies for each expert, CoRDE restricts expert parameters exclusively to LoRA weights on a shared trunk, effectively bypassing the parameter explosion dilemma.

\begin{figure}[t]
    \centering
    \includegraphics[width=\linewidth]{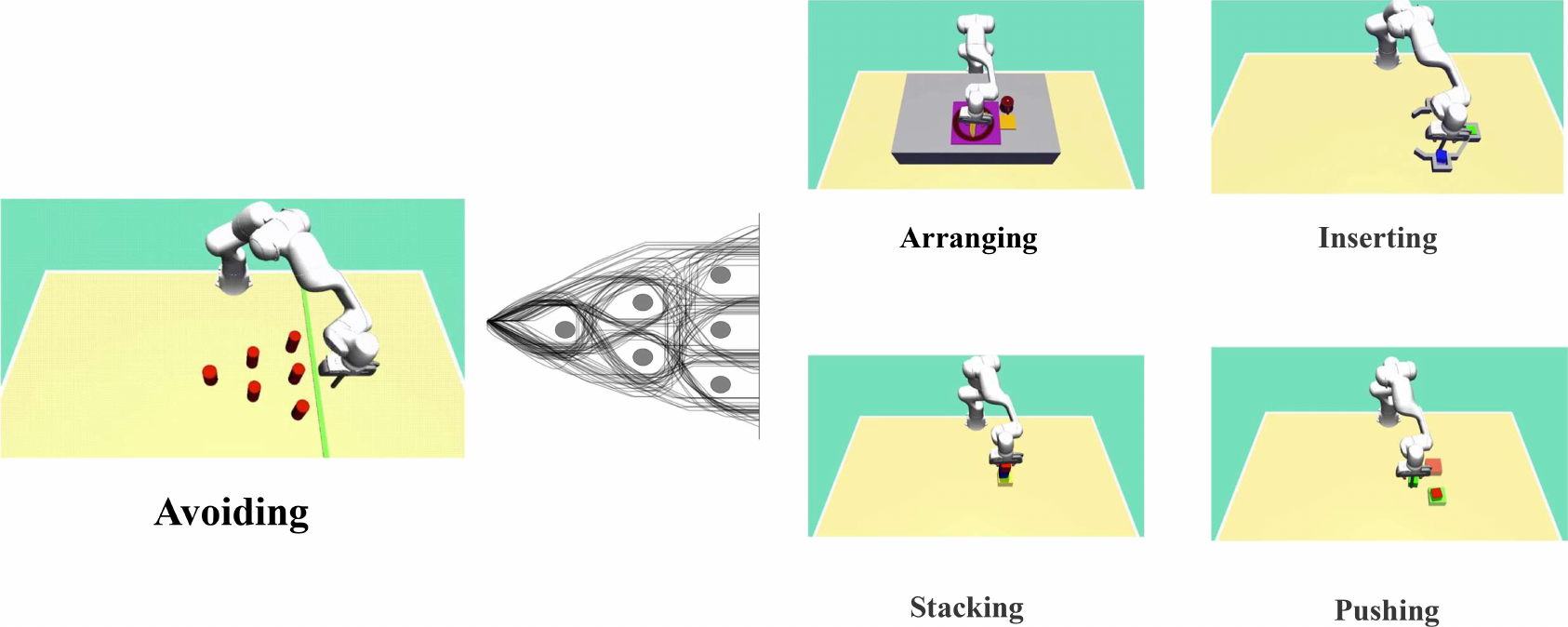}
    \caption{Visualization of the D3IL benchmark tasks used in our evaluation.
    D3IL is designed to stress-test imitation learning under highly multi-modal human demonstrations.
    Left: \textit{Avoiding} admits many distinct successful solutions, illustrated by diverse trajectories.
    Right: representative scenes of \textit{Arranging}, \textit{Inserting}, \textit{Stacking}, and \textit{Pushing}.}
    \label{fig:d3il_tasks_vis}
\end{figure}

\textit{3) Qualitative Visualization of Multimodal demonstration:}
Beyond task success, D3IL explicitly targets \emph{behavioral diversity} by providing human demonstrations with multiple distinct solution modes per task, making it a stringent benchmark for diagnosing mode collapse in imitation learning.
For instance, the \textit{Avoiding} task contains a large set of qualitatively different collision-free strategies, while tasks such as \textit{Aligning}/\textit{Pushing}/\textit{Sorting}/\textit{Stacking} exhibit multi-modality induced by alternative contact choices, object-goal assignments, and different feasible interaction sequences~\cite{jia2024towards}.
Figure~\ref{fig:d3il_tasks_vis} visualizes the task suite and highlights this inherent multi-modality: even within a fixed task specification, the demonstrations form multiple valid trajectory families rather than a single canonical path.
This property makes D3IL particularly suitable for evaluating whether a distilled policy preserves the diversity of human behaviors instead of regressing to an averaged, deterministic solution, a failure mode also emphasized in diffusion-to-MoE distillation studies~\cite{zhou2024variational}.

\subsection{Ablation Study: What Stabilizes Routing and Preserves Diversity?}
\label{subsec:ablation}

We ablate the key design choices in CoRDE that are responsible for \emph{concept--expert decoupling} and
\emph{stable responsibility inference}. All variants follow the same training protocol as the main experiments
and are built upon \textbf{CoRDE (AutoCGP encoder)} as the default setting, so that the effect of each removed
component can be isolated without introducing human concept labels.
To evaluate each design choice from complementary perspectives, we report
\textbf{LIBERO Mean Success} for task performance,
\textbf{D3IL Task Success} and \textbf{Task Entropy} for behavioral diversity,
and a set of routing diagnostics that explicitly distinguish \emph{structured specialization} from \emph{expert collapse}.

\vspace{0.25em}
\noindent\textbf{Routing diagnostics.}
Let $z\in\{1,\dots,K\}$ denote the expert index and $o$ the observation. We compute:
\begin{align}
H_{\text{route}} &\triangleq \mathbb{E}_{o}\!\left[H\!\left(q_{\psi}(z\mid o)\right)\right], \\
H_{\text{post}} &\triangleq \mathbb{E}_{(o,a_0)}\!\left[H\!\left(q^{*}(z\mid o,a_0)\right)\right], \\
u(z) &\triangleq \mathbb{E}_{o}\!\left[q_{\psi}(z\mid o)\right], \qquad
\mathrm{PPL} \triangleq \exp\!\left(H(u)\right),
\end{align}
where $H(p)=-\sum_z p(z)\log p(z)$. Intuitively, $H_{\text{route}}$ measures routing uncertainty at inference,
$H_{\text{post}}$ measures responsibility uncertainty in the E-step, and $\mathrm{PPL}$ measures global expert usage
(collapse typically manifests as $\mathrm{PPL}\!\approx\!1$).

\newcommand{\na}{--} 

\begin{table*}[t]
\centering
\caption{Ablation study. Each variant removes one mechanism from the full CoRDE pipeline. 
We report LIBERO mean success for task performance, D3IL success/entropy for behavioral diversity, and routing diagnostics ($H_{\text{route}}$, $H_{\text{post}}$, $\mathrm{PPL}$) to distinguish structured specialization from expert collapse.}
\label{tab:ablation}
\footnotesize
\setlength{\tabcolsep}{5.0pt}
\renewcommand{\arraystretch}{1.15}

\begin{tabular}{lcccccc}
\toprule
\multirow{2}{*}{\textbf{Variant}} 
& \multicolumn{1}{c}{\textbf{LIBERO} } 
& \multicolumn{2}{c}{\textbf{D3IL} } 
& \multicolumn{3}{c}{\textbf{Routing / Collapse Diagnostics}} \\
\cmidrule(lr){2-2}\cmidrule(lr){3-4}\cmidrule(lr){5-7}
& \makecell{\textbf{Mean Success} \\ $\uparrow$}
& \makecell{\textbf{Success} \\ $\uparrow$}
& \makecell{\textbf{Task Entropy} \\ $\uparrow$}
& \makecell{$H_{\text{route}}$ \\ $\downarrow$}
& \makecell{$H_{\text{post}}$ \\ $\downarrow$}
& \makecell{$\mathrm{PPL}$ \\ $\uparrow$} \\
\midrule

\textbf{Full CoRDE (AutoCGP)} 
& 0.87 & 0.85 & 0.58 & 0.28 & 0.19 & 3.6 \\

w/o $P_{CE}$ (identity mapping) 
& 0.76 & 0.74 & 0.37 & 0.42 & 0.28 & 2.3\\

w/o adaptive $\lambda(o)$ (fixed $\lambda$) 
& 0.82 & 0.67 & 0.52 & 0.18 & 0.14 & 1.4 \\

Evidence-only (Distill-MoE; $\lambda(o)\!=\!0$) 
& 0.79 & 0.76 & 0.32& 0.33 & 0.26 & 2.9 \\

Router w/o responsibility supervision 
& 0.62 & 0.59 & 0.37 & 0.12 & 0.33 & 1.1 \\

\bottomrule
\end{tabular}
\end{table*}

\vspace{0.25em}
\noindent\textbf{How to read Table~\ref{tab:ablation}.}
(i) \textbf{Removing $P_{CE}$} tests whether concept--expert decoupling is necessary.
Without the soft concept-to-expert alignment, the concept space is forced to bind to the expert space too rigidly,
which typically weakens semantic consistency, reduces robustness, and degrades routing diagnostics.
(ii) \textbf{Fixing $\lambda(o)$} tests whether uncertainty-aware annealed fusion is necessary.
A large fixed $\lambda$ over-trusts the semantic prior under concept ambiguity and can inject incorrect structural bias,
whereas a small fixed $\lambda$ degenerates toward evidence-only routing and fails to exploit semantic structure.
(iii) \textbf{Evidence-only distillation} provides a strong lower bound in which responsibilities are driven solely by behavioral evidence.
This variant can retain part of the task performance, but usually yields higher routing uncertainty,
weaker expert structure, and lower behavioral diversity than full CoRDE.
(iv) \textbf{Removing responsibility supervision from the router} tests whether routing can be stably learned without fitting the inferred posterior.
In this case, expert usage typically becomes more imbalanced and the collapse risk increases,
supporting our central claim that CoRDE stabilizes routing by supervising the router with
\emph{inference-derived responsibilities} rather than task gradients.


\section{Conclusion}
We presented CoRDE (Concept-prior Routed Diffusion Experts), a concept-guided diffusion-to-experts framework for robust and interpretable multi-task robot manipulation. CoRDE injects a semantic structural prior from a frozen concept encoder and a learnable concept-to-expert soft mapping, and combines this prior with action evidence through a responsibility inference mechanism to obtain stable expert assignments without relying on task-gradient-driven routing. To achieve scalable deployment, we build a parameter-efficient expert pool by freezing a shared diffusion backbone and training only per-expert LoRA adapters with responsibility-weighted score matching, which preserves both action fidelity and multi-modal behavioral diversity. Experiments on LIBERO and D3IL demonstrate that CoRDE improves task success, maintains higher behavioral diversity, and significantly accelerates inference compared to strong diffusion and distillation baselines. Future work will explore richer open-vocabulary concepts, online adaptation under distribution shifts, and broader real-world validation.

\addtolength{\textheight}{-12cm}   









\bibliographystyle{IEEEtran}
\bibliography{references}

\end{document}